\documentclass[11pt,a4paper]{article}
\usepackage[hyperref]{emnlp-ijcnlp-2019}
\usepackage{times}
\usepackage{latexsym}
\usepackage{amsmath}
\usepackage{amsfonts}
\usepackage{subcaption}

\usepackage{url}
\usepackage{booktabs}
\usepackage{multirow}
\usepackage{wrapfig}
\usepackage{placeins}
\usepackage{tkz-fct}
\usepackage[ngerman]{babel}

\usetikzlibrary{positioning, fit, arrows.meta}

\tikzstyle{neuronSmall}=[rectangle,
thick,
line width=0.5mm,
minimum size=1.2cm,
draw=black,
text centered,
minimum height=3em,
fill=white!20]

\tikzstyle{neuron}=[circle,
thick,
line width=0.5mm,
minimum size=1.3cm,
draw=black,
text centered,
minimum height=2em,
fill=white!20]
\tikzstyle{neuronSmall}=[circle,
thick,
line width=0.5mm,
minimum size=0.7cm,
draw=black,
text centered,
minimum height=2em,
fill=white!20]
\tikzstyle{neuronNoBorder}=[circle,
thick,
minimum size=1.2cm,
draw=none,
fill=white!20]

\newcommand{\bleu}{\textsc{Bleu}}
\newcommand{\ter}{\textsc{Ter}}

\newcommand*{\fcur}{\mathbf{f}_\textrm{cur}}
\newcommand*{\fpre}{\mathbf{f}_\textrm{pre}}
\newcommand*{\hcur}{H_\textrm{cur}}
\newcommand*{\hpre}{H_\textrm{pre}}

\aclfinalcopy 

\title{When and Why is Document-level Context Useful\\in Neural Machine Translation?}

\author{Yunsu Kim \hspace{9pt} Duc Thanh Tran \hspace{9pt} Hermann Ney\\
  Human Language Technology and Pattern Recognition Group \\
  RWTH Aachen University, Aachen, Germany \\
  {\tt \{surname\}@cs.rwth-aachen.de} \\}

\date{}

\begin{document}
\maketitle
\begin{abstract}
  Document-level context has received lots of attention for compensating neural machine translation (NMT) of isolated sentences.
  However, recent advances in document-level NMT focus on sophisticated integration of the context, explaining its improvement with only a few selected examples or targeted test sets.
  We extensively quantify the causes of improvements by a document-level model in general test sets, clarifying the limit of the usefulness of document-level context in NMT.
  We show that most of the improvements are not interpretable as utilizing the context.
  We also show that a minimal encoding is sufficient for the context modeling and very long context is not helpful for NMT.\\
\end{abstract}

\section{Introduction}
\label{sec:introduction}

Neural machine translation (NMT) \cite{bahdanau2015neural,vaswani2017attention} has been originally developed to work sentence by sentence. Recently, it has been claimed that sentence-level NMT generates document-level errors, e.g. wrong coreference of pronouns/articles or inconsistent translations throughout a document \cite{guillou2018pronoun,laubli2018has}.

A lot of research addresses these problems by feeding surrounding context sentences as additional inputs to an NMT model.
Modeling of the context is usually done with fully-fledged NMT encoders with extensions to consider complex relations between sentences \cite{bawden2018evaluating,voita2018context,zhang2018improving,miculicich2018document,maruf2019selective}.
Despite the high overhead in modeling, translation metric scores (e.g. {\bleu}) are often only marginally improved, leaving the evaluation to artificial tests targeted for pronoun resolution \cite{jean2017does,tiedemann2017neural,bawden2018evaluating,voita2018context,voita2019when}.
Even if the metric score gets significantly better, the improvement is limited to specific datasets or explained with only a few examples \cite{tu2018learning,maruf2018document,kuang2018fusing,cao2018encoding,zhang2018improving,maruf2019selective}.

This paper systematically investigates when and why document-level context improves NMT, asking the following research questions:
\begin{itemize}\itemsep0em
    \item In general, how often is the context utilized in an interpretable way, e.g. coreference?
    \item Is there any other (non-linguistic) cause of improvements by document-level models?
    \item Which part of a context sentence is actually meaningful for the improvement?
    \item Is a long-range context, e.g. in ten consecutive sentences, still useful?
    \item How much modeling power is necessary for the improvements?
\end{itemize}

To answer these questions, we conduct an extensive qualitative analysis on non-targeted test sets.
According to the analysis, we use only the important parts of the surrounding sentences to facilitate the integration of long-range contexts.
We also compare different architectures for the context modeling and check sufficient model complexity for a significant improvement.

Our results show that the improvement in {\bleu} is mostly from a non-linguistic factor: regularization by reserving parameters for context inputs.
We also verify that very long context is indeed not helpful for NMT, and a full encoder stack is not necessary for the improved performance.

\section{Document-level NMT}
\label{sec:doc-level}

In this section, we review the existing document-level approaches for NMT and describe our strategies to filter out uninteresting words in the context input.
We illustrate with an example of including one previous source sentence as the document-level context, which can be easily generalized also to other context inputs such as target hypotheses \cite{agrawal2018contextual,bawden2018evaluating,voita2019when} or decoder states \cite{tu2018learning,maruf2018document,miculicich2018document}.

For the notations, we denote a source sentence by $\mathbf{f}$ and its encoded representations by $H$.
A subscript distinguishes the previous (pre) and current (cur) sentences.
$e_i$ indicates a target token to be predicted at position $i$, and $e_1^{i-1}$ are already predicted tokens in previous positions.
$Z$ denotes encoded representations of a partial target sequence.

\subsection{Single-Encoder Approach}
\label{ssec:single-enc}

The simplest method to include context in NMT is to just modify the input, i.e. concatenate surrounding sentences to the current one and put the extended sentence in a normal sentence-to-sentence model \cite{tiedemann2017neural,agrawal2018contextual}.
A special token is inserted between context and current sentences to mark sentence boundaries (e.g. \texttt{\_BREAK\_}).

\begin{figure}[!ht]
	\centering
	\scalebox{0.5}{
		\begin{tikzpicture}[auto]
		\tikzstyle{Encoder}=[rectangle, draw=black, thin, fill=purple!40, minimum width=4.5cm, minimum height=1.5cm];
		\tikzstyle{Multihead}=[rounded rectangle, draw=black, thin, fill=orange!40, minimum width=4.5cm, minimum height=1cm];
		\tikzstyle{FFN}=[rounded rectangle, draw=black, thin, fill=blue!20, minimum width=4.5cm, minimum height=1cm];
		\tikzstyle{Operation} = [inner sep=0pt, thin];
		\tikzstyle{Arrow}=[->, >=latex', shorten >=2pt, very thick];
		
		\tikzstyle{Decoder}=[rectangle, draw=black, thin, fill=teal!40, minimum width=4.5cm, minimum height=8.5cm];
		
		\node[Encoder] (src_enc) at (0,0.5) {};
		\node[above right] at (src_enc.south west) {\LARGE Encoder$_{\textrm{pre+cur}}$};
		
		\node[Decoder] (decoder) at (8, 4) {};
		\node[above right] at (decoder.south west) {\LARGE Decoder};
		\node[below left] at (decoder.north east) {\LARGE $\times N$};
		\node[Multihead] (enc_dec_attention) at (8, 6) {\LARGE Attention};

		\node (src_input) at (0,-2.0)
		{\LARGE $\fpre$ {\large \texttt{\_BREAK\_}} $\fcur$};
		\node (tar_input) at (9, -2.0)
		{\LARGE $e_1^{i-1}$};
		\node (src_enc_repr) at (0,2.5)
		{\LARGE $H_\textrm{pre+cur}$};
		\node (dec_self_attended) at (9,2)
		{\LARGE $Z$};
		
		\node (out) at (8,10.0)
		{\LARGE $p\left(e_i|e_1^{i-1}, \fcur, \fpre\right)$};

		\draw[Arrow] (src_input.north) -- +(0,1);
		\draw[Arrow] (tar_input.north) -- +(0,0.9);
		\draw[Arrow] (src_enc.north) -- +(0,0.8);
		
		\draw[Arrow] (src_enc_repr.east) to [out=2,in=270] (enc_dec_attention.205);
		\draw[Arrow] (src_enc_repr.east) to [out=4,in=270] (enc_dec_attention.265);
		
		\draw[Arrow, dashed] (9,0) -- (dec_self_attended.south);
		\draw[Arrow] (dec_self_attended.north) -- +(0,3.1);
		\draw[Arrow, dashed] (enc_dec_attention.north) -- +(0,1.5);
		\draw[Arrow] (decoder.north) -- (out.south);
		
		\end{tikzpicture}
	}
	\caption{Single-encoder approach.}
	\label{fig:docnmt:concat}
\end{figure}
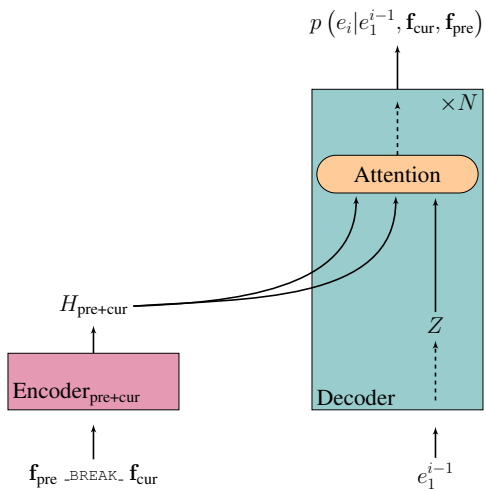

Figure \ref{fig:docnmt:concat} depicts this approach.
Here, a single encoder processes the context and current sentences together as one long input.
This requires no change in the model architecture but worsens a fundamental problem of NMT: translating long inputs \cite{koehn2017six}.
Apart from the data scarcity of a higher-dimensional input space, it is difficult to optimize the attention component to the long spans \cite{sukhbaatar2019adaptive}. 

\subsection{Multi-Encoder Approach}
\label{ssec:multi-enc}

Alternatively, multi-encoder approaches encode each additional sentence separately.
The model learns representations solely of the context sentences which are then integrated into the baseline model architecture.
This tackles the integration of additional sentences on the architecture level, in contrast to the single-encoder approach.
In the following, we describe two methods of integrating the encoded context sentences.
The descriptions below do not depend on specific types of context encoding; one can use recurrent or self-attentive encoders with a variable number of layers, or just word embeddings without any hidden layers on top of them (Section \ref{ssec:architecture}).

\subsubsection{Integration Outside the Decoder}
\label{ssec:multi-enc:outside}

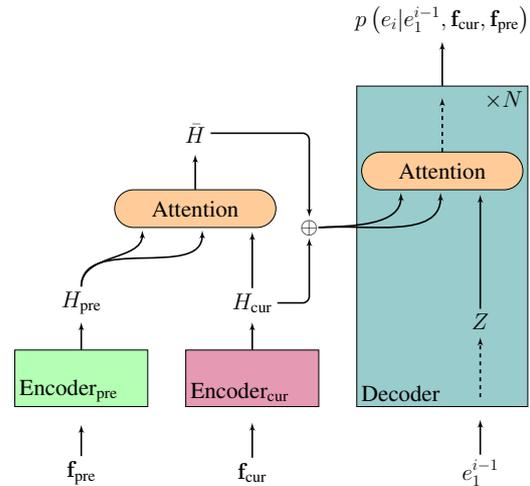
\begin{figure}[!h]
\centering
\scalebox{0.5}{
\begin{tikzpicture}[auto]
	\tikzstyle{Encoder}=[rectangle, draw=black, thin, fill=purple!40, minimum width=3.5cm, minimum height=1.5cm];
	\tikzstyle{Multihead}=[rounded rectangle, draw=black, thin, fill=orange!40, minimum width=4.5cm, minimum height=1cm];
	\tikzstyle{FFN}=[rounded rectangle, draw=black, thin, fill=blue!20, minimum width=4.5cm, minimum height=1cm];
	\tikzstyle{Operation} = [inner sep=0pt, thin];
	\tikzstyle{Arrow}=[->, >=latex', shorten >=2pt, very thick];
	
	\tikzstyle{Decoder}=[rectangle, draw=black, thin, fill=teal!40, minimum width=4.5cm, minimum height=8.5cm];

	\node[Encoder] (src_enc) at (0.5,0.5) {}; 
	\node[above right] at (src_enc.south west) {\LARGE Encoder$_{\textrm{cur}}$};
	\node[Encoder, fill=green!30] at (-4,0.5) (ctx_enc) {};
	\node[above right] at (ctx_enc.south west) {\LARGE Encoder$_{\textrm{pre}}$};
	\node[Multihead] (multihead) at (-1,5) {\LARGE Attention};
	\node[Operation] (combination) at (2,4.5) {\LARGE $\oplus$};
	
	
	\node[Decoder] (decoder) at (5.5, 4) {};
	\node[above right] at (decoder.south west) {\LARGE Decoder};
	\node[below left] at (decoder.north east) {\LARGE $\times N$};

	\node[Multihead] (enc_dec_attention) at (5.5, 6) {\LARGE Attention};

	\node (src_input) at (0.5,-2)
		{\LARGE $\fcur$};
\node (ctx_input) at (-4, -2){\LARGE $\fpre$};
	\node (tar_input) at (6.5, -2)
		{\LARGE $e_1^{i-1}$};
	\node (src_enc_repr) at (0.5,2.5)
		{\LARGE $\hcur$};
	\node(ctx_enc_repr) at (-4,2.5) {\LARGE $\hpre$};
	\node (attended) at (-1,7) {\LARGE $\bar{H}$};
	\node (dec_self_attended) at (6.5,2)
		{\LARGE $Z$};
	\node (out) at (5.5,10)
		{\LARGE $p\left(e_i|e_1^{i-1}, \fcur, \fpre\right)$};

	\draw[Arrow] (src_input.north) -- +(0,1);
	\draw[Arrow] (ctx_input.north) -- +(0,0.9);
	\draw[Arrow] (tar_input.north) -- +(0,1);
	\draw[Arrow] (src_enc.north) -- +(0,0.8);
	\draw[Arrow] (ctx_enc.north) -- +(0,0.8);
	\draw[Arrow] (ctx_enc_repr.north) to [out=90,in=270] (multihead.200);
	\draw[Arrow] (ctx_enc_repr.north) to [out=90,in=270] (multihead.290);
	\draw[Arrow] (src_enc_repr.north) -- (0.5,4.45);
	\draw[Arrow] (multihead.north) -- +(0,1);
	\draw[Arrow,rounded corners=5pt] (attended.east) -| (combination);
	\draw[Arrow,rounded corners=5pt] (src_enc_repr.east) -| (combination);
	
	\draw[Arrow, dashed] (6.5,0) -- (dec_self_attended.south);
	\draw[Arrow] (dec_self_attended.north) -- +(0,3.1);
	\draw[Arrow] (combination.east) to [out=2,in=270] (enc_dec_attention.205);
	\draw[Arrow] (combination.east) to [out=4,in=270] (enc_dec_attention.265);
	\draw[Arrow, dashed] (enc_dec_attention.north) -- +(0,1.5);
	\draw[Arrow] (decoder.north) -- (out.south);
	
\end{tikzpicture}
}
\caption{Multi-encoder approach integrating context outside the decoder.}
\label{fig:docnmt:outside}\vspace{-0.5em}
\end{figure}

The first method combines encoder representation of all input sentences before being fed to the decoder \cite{maruf2018document,voita2018context,miculicich2018document,zhang2018improving,maruf2019selective}.
It attends from the representations of the current sentence ($H_\textrm{cur}$) to those of the previous sentence ($H_\textrm{pre}$), yielding $\bar{H}$.
Afterwards, a linear interpolation with gating is applied:
\begin{equation}\label{eq:gating}
	g\bar{H} + (1 - g)H_\textrm{cur}
\end{equation}
where $g=\sigma\left(W_g\left[\bar{H};H_\textrm{cur}\right]+b_g\right)$ is gating activation and $W_g,b_g$ are learnable parameters.
This type of integration is depicted in Figure \ref{fig:docnmt:outside}.
By using such a gating mechanism, the model is capable of learning how much additional context information shall be included.

\subsubsection{Integration Inside the Decoder}
\label{ssec:multi-enc:inside}

Another method integrates the context inside the decoder; the partial target history $e_1^{i-1}$ is available during the integration.
Here, using the (encoded) target history as a query, the decoder attends directly to the context representations.
It also has the original attention to the current sentence.
Depending on the order of these two attention components, this type of integration has two variants.

\vspace{1em}\noindent\textbf{Sequential Attentions}\hspace{0.3cm}
The first variant is stacking the two attention components, with the output of one component being the query of another \cite{tu2018learning,zhang2018improving}.

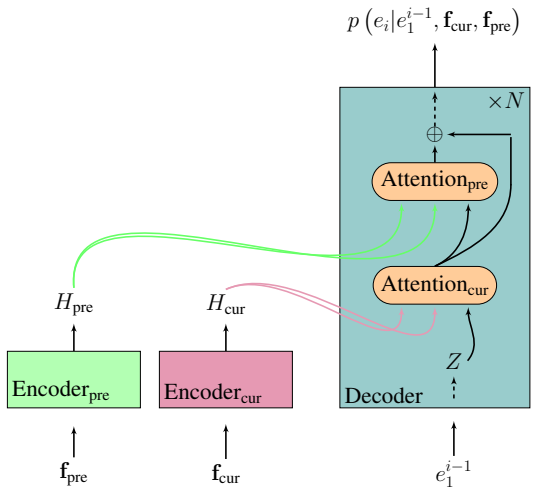
\begin{figure}[!h]
\centering
\scalebox{0.5}{
\begin{tikzpicture}[auto]
	\tikzstyle{Encoder}=[rectangle, draw=black, thin, fill=purple!40, minimum width=3.5cm, minimum height=1.5cm];
	\tikzstyle{Multihead}=[rounded rectangle, draw=black, thin, fill=orange!40, minimum width=3.0cm, minimum height=1cm];
	\tikzstyle{FFN}=[rounded rectangle, draw=black, thin, fill=blue!20, minimum width=4.5cm, minimum height=1cm];
	\tikzstyle{Operation} = [inner sep=0pt, thin];
	\tikzstyle{Arrow}=[->, >=latex', shorten >=2pt, very thick];
	
	\tikzstyle{Decoder}=[rectangle, draw=black, thin, fill=teal!40, minimum width=5cm, minimum height=8.5cm];

	\node[Encoder] (src_enc) at (0.5,0.5) {};
	\node[above right] at (src_enc.south west) {\LARGE Encoder$_{\textrm{cur}}$};
	\node[Encoder, fill=green!30] at (-3.5,0.5) (ctx_enc) {};
	\node[above right] at (ctx_enc.south west) {\LARGE Encoder$_{\textrm{pre}}$};
	
	\node[Decoder] (decoder) at (6, 4) {};
	\node[above right] at (decoder.south west) {\LARGE Decoder};
	\node[below left] at (decoder.north east) {\LARGE $\times N$};
	
	\node[Multihead] (enc_dec_attention) at (6, 3) {\LARGE Attention$_\textrm{cur}$};
	\node[Multihead] (enc_ctx_dec_attention) at (6, 5.75) {\LARGE Attention$_\textrm{pre}$};
	
	\node[Operation] (combination) at (6,7) {\LARGE$\oplus$};

	\node (src_input) at (0.5,-2)
		{\LARGE $\fcur$};
	\node (ctx_input) at (-3.5,-2)
		{\LARGE $\fpre$};
	\node (tar_input) at (6.5, -2)
		{\LARGE $e_1^{i-1}$};
	\node (src_enc_repr) at (0.5,2.5)
		{\LARGE $\hcur$};
	\node (ctx_enc_repr) at (-3.5,2.5)
		{\LARGE $\hpre$};
	\node (dec_self_attended) at (6.5,1)
		{\LARGE $Z$};
	
	\node (out) at (6,10)
		{\LARGE $p\left(e_i|e_1^{i-1}, \fcur, \fpre\right)$};
	
	\draw[Arrow] (src_input.north) -- +(0,1);
	\draw[Arrow] (ctx_input.north) -- +(0,0.9);
	\draw[Arrow] (tar_input.north) -- +(0,1);
	\draw[Arrow] (src_enc.north) -- +(0,0.8);
	\draw[Arrow] (ctx_enc.north) -- +(0,0.8);
	\draw[Arrow, purple!40] (src_enc_repr.north) to [out=25,in=270] (enc_dec_attention.270);
	\draw[Arrow, purple!40] (src_enc_repr.north) to [out=25,in=270] (enc_dec_attention.210);
	\draw[Arrow, green!60] (ctx_enc_repr.north) to [out=95,in=270] (enc_ctx_dec_attention.270);
	\draw[Arrow, green!60] (ctx_enc_repr.north) to [out=95,in=270] (enc_ctx_dec_attention.210);
	\draw[Arrow,rounded corners=5pt] (enc_ctx_dec_attention.north) -- (combination.south);
	
	\draw[Arrow] (enc_dec_attention.north) to [out=25,in=270] (enc_ctx_dec_attention.330);
	\draw[>=latex', shorten >=2pt, very thick] (enc_dec_attention.north) to [out=25,in=270] (8,5.75);
	\draw[>=latex', shorten >=2pt, very thick] (8,5.65) -- (8,7);
	\draw[Arrow] (8,7) -- (combination.east);
	
	\draw[Arrow, dashed] (6.5,0) -- (dec_self_attended.south);
	\draw[Arrow] (dec_self_attended.east) to [out=50,in=270] (enc_dec_attention.330);
	\draw[Arrow, dashed] (combination.north) -- +(0,1);
	\draw[Arrow] (decoder.north) -- (out.south);
	
\end{tikzpicture}
}
\caption{Multi-encoder approach integrating context inside the decoder with sequential attentions.}
\label{fig:decoder_attention_parallel:seq}
\end{figure}

Figure \ref{fig:decoder_attention_parallel:seq} shows the case when the current sentence is attended by the decoder first, which is then used to attend to the context sentence.
This refines the regular attention to the current source sentence with additional context information.
The order of the attention components may be switched.
To block signals of potentially unimportant context information, a gating mechanism can be employed between the regular and context attention outputs like Section \ref{ssec:multi-enc:outside}.
\vspace{0.5em}

\begin{figure}[!h]
\centering
\scalebox{0.5}{
\begin{tikzpicture}[auto]
	\tikzstyle{Encoder}=[rectangle, draw=black, thin, fill=purple!40, minimum width=3.5cm, minimum height=1.5cm];
	\tikzstyle{Multihead}=[rounded rectangle, draw=black, thin, fill=orange!40, minimum width=3.0cm, minimum height=1cm];
	\tikzstyle{FFN}=[rounded rectangle, draw=black, thin, fill=blue!20, minimum width=4.5cm, minimum height=1cm];
	\tikzstyle{Operation} = [inner sep=0pt, thin];
	\tikzstyle{Arrow}=[->, >=latex', shorten >=2pt, very thick];
	
	\tikzstyle{Decoder}=[rectangle, draw=black, thin, fill=teal!40, minimum width=7cm, minimum height=8cm];

	\node[Encoder] (src_enc) at (0,0.7) {}; 
	\node[above right] at (src_enc.south west) {\LARGE Encoder$_{\textrm{cur}}$};
	\node[Encoder, fill=green!30] at (-4,0.7) (ctx_enc) {};
	\node[above right] at (ctx_enc.south west) {\LARGE Encoder$_{\textrm{pre}}$};
	
	\node[Decoder] (decoder) at (6, 4) {};
	\node[above right] at (decoder.south west) {\LARGE Decoder};
	\node[below left] at (decoder.north east) {\LARGE $\times N$};

	\node[Multihead] (enc_dec_attention) at (7.7, 4.75) {\LARGE Attention$_\textrm{cur}$};
	\node[Multihead] (enc_ctx_dec_attention) at (4.3, 4.75) {\LARGE Attention$_\textrm{pre}$};
	
	\node[Operation] (combination) at (6,6.5) {\LARGE $\oplus$};

	\node (src_input) at (0,-2)
		{\LARGE $\fcur$};
	\node (ctx_input) at (-4,-2)
		{\LARGE $\fpre$};
	\node (tar_input) at (6, -2)
		{\LARGE $e_1^{i-1}$};
	\node (src_enc_repr) at (0,2.5)
		{\LARGE $\hcur$};
	\node (ctx_enc_repr) at (-4,2.5)
		{\LARGE $\hpre$};
	\node (dec_self_attended) at (6,1.75)
		{\LARGE $Z$};
	
	\node (out) at (6,10)
		{\LARGE $p\left(e_i|e_1^{i-1}, \fcur, \fpre\right)$};
	
	\draw[Arrow] (src_input.north) -- +(0,1);
	\draw[Arrow] (ctx_input.north) -- +(0,0.9);
	\draw[Arrow] (tar_input.north) -- +(0,1);
	\draw[Arrow] (src_enc.north) -- +(0,0.8);
	\draw[Arrow] (ctx_enc.north) -- +(0,0.8);
	\draw[Arrow, purple!40] (src_enc_repr.north) to [out=25,in=270] (enc_dec_attention.270);
	\draw[Arrow, purple!40] (src_enc_repr.north) to [out=25,in=270] (enc_dec_attention.210);
	\draw[Arrow, green!60] (ctx_enc_repr.north) to [out=25,in=270] (enc_ctx_dec_attention.270);
	\draw[Arrow, green!60] (ctx_enc_repr.north) to [out=25,in=270] (enc_ctx_dec_attention.210);
	\draw[Arrow,rounded corners=5pt] (enc_ctx_dec_attention.north) |- (combination.west);
	\draw[Arrow,rounded corners=5pt] (enc_dec_attention.north) |- (combination);
	
	\draw[Arrow, dashed] (6,0.2) -- (dec_self_attended.south);
	\draw[Arrow] (dec_self_attended.east) to [out=25,in=270] (enc_dec_attention.330);
	\draw[Arrow] (dec_self_attended.west) to [out=180,in=270] (enc_ctx_dec_attention.330);
	\draw[Arrow, dashed] (combination.north) -- +(0,1.2);
	\draw[Arrow] (decoder.north) -- (out.south);
	
\end{tikzpicture}
}
\caption{Multi-encoder approach integrating context inside the decoder with parallel attentions.}
\label{fig:decoder_attention_parallel}\vspace{-0.3em}
\end{figure}
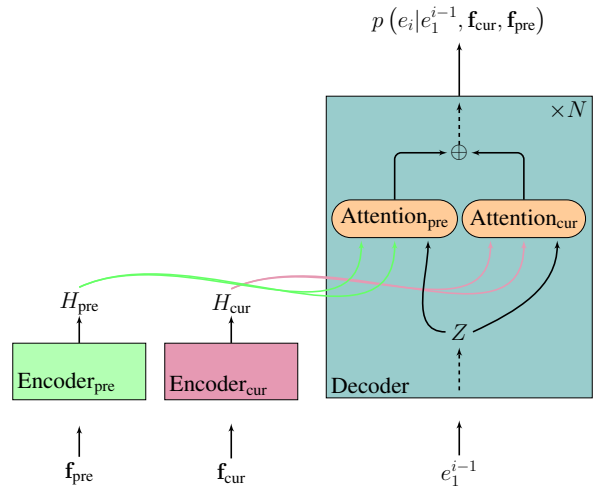

\vspace{0.7em}\noindent\textbf{Parallel Attentions}\hspace{0.3cm}
Figure \ref{fig:decoder_attention_parallel} shows the case when performing the two attention operations in parallel and combining them with a gating afterwards \cite{jean2017does,cao2018encoding,kuang2018fusing,bawden2018evaluating,stojanovski2018coreference}.
This method relates document-level context to the target history independently of the current source sentence, and lets the decoding computation faster.

\noindent For each category above, we have described a common architecture shared by previous works in that category.
There are slight variations but they do not diverge much from our descriptions.

\subsection{Filtering of Words in the Context}
\label{ssec:summary-theory}

\begin{table*}[!ht]
	\begin{center}
		\begin{tabular}{p{3.5cm}|p{11cm}}
		\toprule
			Original source& in recent years, I correctly foresaw that, in the absence of stronger fiscal stimulus (which was not forthcoming in either Europe or the United States), recovery from the Great Recession of 2008 would be slow.\\
			\midrule\midrule
			Remove stopwords & recent years, I correctly foresaw  absence stronger fiscal stimulus (forthcoming Europe United States), recovery Great Recession 2008 slow. \\
			\midrule
			Remove most frequent words & recent correctly foresaw absence stronger fiscal stimulus forthcoming either States recovery Great Recession 2008 slow \\
			\midrule
			Retain named entities &  recent years Europe the United States the Great Recession 2008 \\
			\midrule
			Retain specific POS &  years I foresaw the absence stimulus was forthcoming either Europe or the United States recovery the Great Recession 2008 would be \\
			\bottomrule
		\end{tabular}
		\caption{Examples for filtering of words in the context (News Commentary v14 English$\to$German).}\label{tab:summary-ex}
	\end{center}
\end{table*}

Document-level NMT inherently has heavy computations due to longer inputs and additional processing of context.
However, intuitively, not all of the words in the context are actually useful in translating the current sentence.
For instance, in most literature, the improvements from using document-level context are explained with coreference, which can be resolved with just nouns, articles, and the conjugated words affected by them.

Under the assumption that we do not need the whole context sentence in document-level NMT, we suggest to retain only the context words that are likely to be useful.
This makes the training easier with a smaller input space and less memory requirement.
Concretely, we filter out words in the context sentences according to pre-defined word lists or predicted linguistic tags:
\begin{itemize}\itemsep0em
	\item Remove stopwords using a pre-defined list\footnote{https://github.com/explosion/spaCy}
	\item Remove $n\in\mathbb{N}$ most frequent words
	\item Retain only named entities
	\item Retain only the words with specific parts-of-speech (POS) tags
\end{itemize}
The first method has the same motivation as \newcite{kuang2018modeling} to ignore function words.
The second method aims to keep infrequent words that are domain-specific or containing gender information.
We empirically found that $n=150$ works reasonably well.
For the last two methods, we use the \textsc{Flair}\footnote{https://github.com/zalandoresearch/flair} \cite{akbik2018flair} toolkit.
We exclude the tags that are irrelevant to syntax/semantics of the current sentence.
The detailed lists of retained tags can be found in the appendix.

The filtering is performed on word level in the preprocessing.
When a sentence is completely pruned, we use a special token to denote an empty sentence (e.g. \texttt{\_EMPTY\_}).
Table \ref{tab:summary-ex} gives examples of the filtering.
We can observe that the original sentence is shortened greatly by removing redundant tokens, but the topic information and the important subjects still remain.

\section{Experiments}

We evaluate the document-level approaches in IWSLT 2017 English$\to$Italian\footnote{https://sites.google.com/site/iwsltevaluation2017} and WMT 2018 English$\to$German\footnote{https://www.statmt.org/wmt18/translation-task.html} translation tasks.
We used TED talk or News Commentary v14 dataset as the training data respectively, preprocessed with theMoses tokenizer\footnote{http://www.statmt.org/moses} and byte pair encoding \cite{sennrich2016neural} trained with 32k merge operations jointly for source and target languages.
In all our experiments, one previous source sentence was given as the document-level context.
A special token was inserted at each document boundary, which was also fed as context input when translating sentences around the boundaries.
Detailed corpus statistics are given in Table \ref{tab:corpus}.

All experiments were carried out with \textsc{Sockeye} \cite{hieber2017sockeye}.
We used Adam optimizer \cite{kingma2014adam} with the default parameters.
The learning rate was reduced by 30\% when the perplexity on a validation set was not improving for four checkpoints.
When it did not improve for ten checkpoints, we stopped the training.
Batch size was 3k tokens, where the bucketing was done for a tuple of current/context sentence lengths.
All other settings follow a 6-layer base Transformer model \cite{vaswani2017attention}.

In all our experiments, a sentence-level model was pre-trained and used to initialize document-level models, which was crucial for the performance. 
We also shared the source word embeddings over the original and context encoders.

\begin{table}[!h]
    \centering
    \begin{tabular}{lcc}
    \toprule
         & \textbf{en-it} & \textbf{en-de}\\
    \midrule
        Running Words & 4.3M & 8.1M\\
        Sentences & 227k & 329k\\
        Documents & 2,045 & 8,891\\
        Document Length (avg. \#sent) & 111 & 37\\
    \bottomrule
    \end{tabular}
    \caption{Training data statistics.}
    \label{tab:corpus}\vspace{-0.5em}
\end{table}

\setlength{\tabcolsep}{3pt}
\begin{table*}[!ht]
    \centering
    \begin{tabular}{cccccccccc}
        \toprule
         & \multicolumn{2}{c}{Context Encoder} & & \multicolumn{2}{c}{\textbf{en-it}} & & \multicolumn{2}{c}{\textbf{en-de}} \\
        \cmidrule{2-3}\cmidrule{5-6}\cmidrule{8-9}
        Approach & Architecture & \#layers & & {\bleu} [\%] & {\ter} [\%] & & {\bleu} [\%] & {\ter} [\%]\\
        \midrule\midrule
        Baseline & $\cdot$ & $\cdot$ & & 31.4 & 56.1 & & 28.9 & 61.8\\
        \midrule
        Single-Encoder & Transformer & 6 & & 31.5 &  57.2&  & 28.9 & 61.4\\
        \midrule
        Multi-Encoder (Out.) & Transformer & 6 & & 31.3 & 56.1 & & 29.1 & 61.4 \\
        \midrule
        Multi-Encoder (Seq.) & Transformer & 6 & & 32.6 & 55.2 & & 29.9 & 60.7\\
        \midrule
        \multirow{4}{*}{Multi-Encoder (Para.)} &  \multirow{3}{*}{Transformer} & 6 & & \textbf{32.7} & \textbf{54.7} & & 30.1 & 60.3\\
        & & 2 & & 32.6 & 55.2 & & 30.2 & 60.5\\
        & & 1 & & 32.2 & 55.8 & & 30.0 & 60.4\\
        \cmidrule{2-9}
        & Word Embedding & $\cdot$ & & 32.5 & 54.8 & & \textbf{30.3} & \textbf{59.9}\\
        \bottomrule
    \end{tabular}
    \caption{Comparison of document-level model architectures and complexity.}
    \label{tab:architecture}
\end{table*}

\subsection{Model Comparison}
\label{ssec:architecture}

\noindent\textbf{Model Architecture}\hspace{0.3cm} Firstly, we compare the performance of existing single-encoder and multi-encoder approaches (Table \ref{tab:architecture}).
For each category of document-level methods (Section \ref{sec:doc-level}), we test one representative architecture  (Figures \ref{fig:docnmt:outside}, \ref{fig:decoder_attention_parallel:seq}, \ref{fig:decoder_attention_parallel}) which encompasses all existing work in that category except slight variations.
The tested methods are equal or closest to:
\begin{itemize}\itemsep0em
    \item Single-Encoder: \newcite{agrawal2018contextual}
    \item Integration outside the decoder: \newcite{voita2018context} without sharing the encoder hidden layers over current/context sentences
    \item Integration inside the decoder
    \begin{itemize}
        \item Sequential attention: Decoder integration of \newcite{zhang2018improving} with the order of attentions (current/context) switched
        \item Parallel attention: Gating version of \newcite{bawden2018evaluating}
    \end{itemize}
\end{itemize}

The training of the single-encoder method was quite unstable. It took about twice as long as other document-level models, yet yielding no improvements, which is consistent with \newcite{kuang2018fusing}.
Longer inputs make the encoder-decoder attention widely scattered and harder to optimize.
We might need larger training data, massive pre-training, and much larger batches to train the single-encoder approach effectively \cite{junczys2019microsoft}; however, these conditions are often not realistic.

For the multi-encoder models, if the context is integrated outside the decoder (``Out.''), it barely improves upon the baseline.
By letting the decoder directly access context sentences with a separate attention component, they all outperform the single-encoder method, improving the sentence-level baseline up to +1.4\% {\bleu} and -1.9\% {\ter}.
Particularly, when attending to current and context sentences in parallel (``Para.''), it provides more flexible and selective information flow from multiple source sentences to the decoder, thus producing better results than the sequential attentions (``Seq.'').
\vspace{0.7em}

\noindent\textbf{Model Complexity}\hspace{0.3cm} In the linguistic sense, surrounding sentences are useful in translating the current sentence mostly by providing case distinctions of nouns or topic information (Section \ref{sec:analysis}).
The sequential relation of tokens in the surrounding sentences is important for neither of them.
Therefore we investigate how many levels of sequential encoding is actually needed for the improvement by the context.
From a 6-layer Transformer encoder, we gradually reduce the model complexity of the context encoder: 2-layer, 1-layer, and only using word embeddings without any sequential encoding.
We remove positional encoding \cite{vaswani2017attention} when we encode only with word embeddings.

The results are shown in the lower part of Table \ref{tab:architecture}.
Context encoding without any sequential modeling (the last row) shows indeed comparable performance to using a full 6-layer encoder.
This simplified encoding eases the memory-intensive document-level training by having 22\% fewer model parameters, which allows us to adopt a larger batch size without accumulating gradients.
For the remainder of this paper, we stick to using the multi-encoder approach with parallel attention components in the decoder and restricting the context encoding to only word embeddings.

\begin{table*}[!ht]
    \centering
    \begin{tabular}{lcccccccc}
        \toprule
         & & \multicolumn{2}{c}{\textbf{en-it}} & & \multicolumn{2}{c}{\textbf{en-de}} \\
        \cmidrule{3-4} \cmidrule{6-7}
        Context sentence & & {\bleu} [\%] & {\ter} [\%] & & {\bleu} [\%] & {\ter} [\%] & & \#tokens\\
        \midrule
        None & & 31.4 & 56.1 & & 28.9 & 61.8 & & -\\
        Full sentence & & 32.5 & 54.8 & & 30.3 & 59.9 & & 100\%\\
        \midrule
        Remove stopwords & & 32.2 & 55.2 & & 30.3 & 59.9 &  & \enspace63\%\\
        Remove most frequent words & & 32.1 & 55.6 & & 30.2 & 60.2 & & \enspace51\%\\
        Retain only named entities & & 32.3 & 55.4 && 30.3 & 60.3 & & \enspace{\bf 13\%}\\
        Retain specific POS & & {\bf 32.5} & {\bf 55.2} & & {\bf 30.4} & {\bf 60.0} & & \enspace59\%\\
        \bottomrule
    \end{tabular}
    \caption{Comparison of context word filtering methods.}
    \label{tab:summarization}
\end{table*}

\begin{figure*}[!ht]
\centering
\begin{subfigure}[t]{0.45\linewidth}
\includegraphics[width=\textwidth]{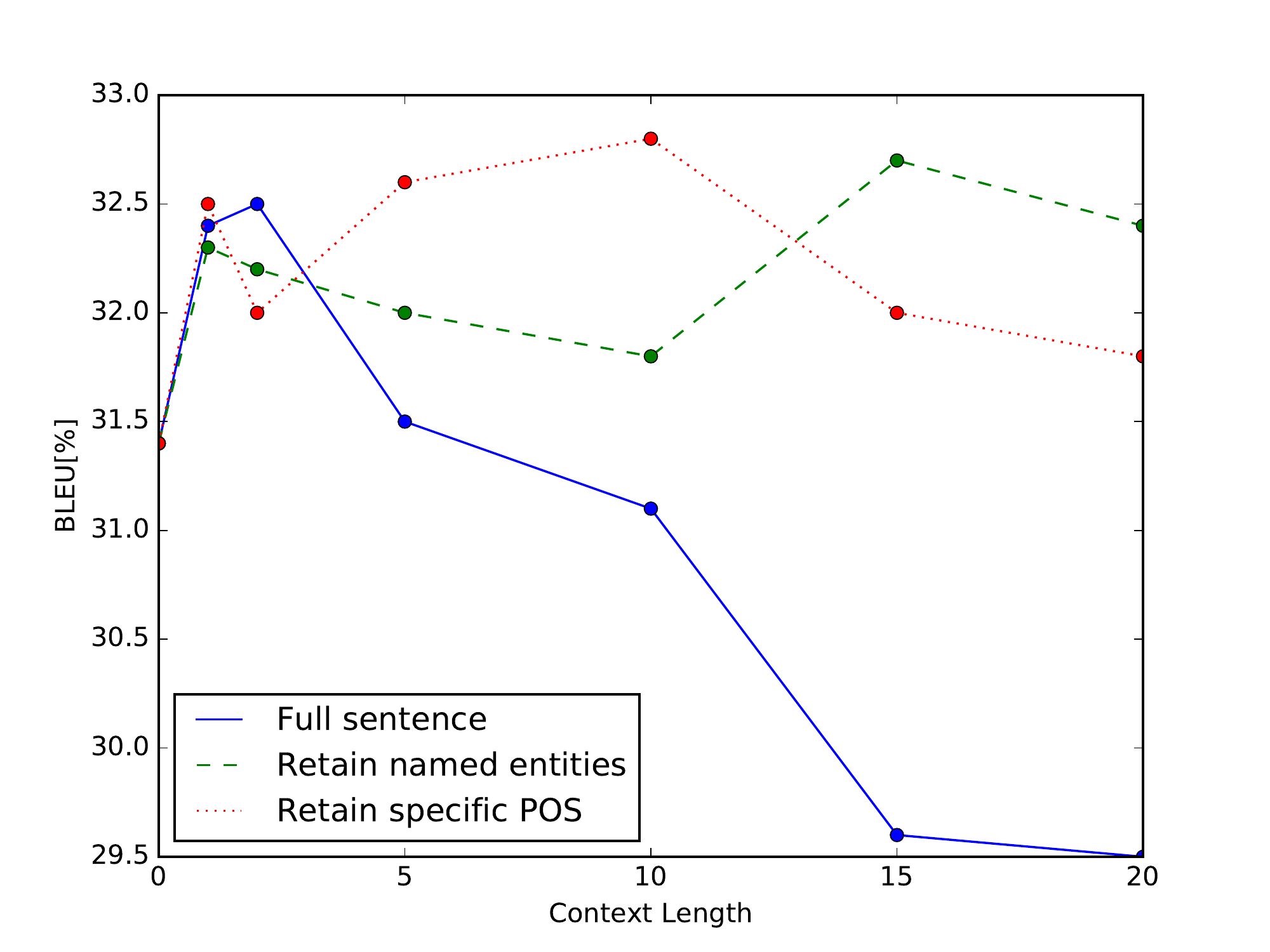}
\caption{IWSLT \textbf{en-it}}
\end{subfigure}
\begin{subfigure}[t]{0.45\linewidth}
\includegraphics[width=\textwidth]{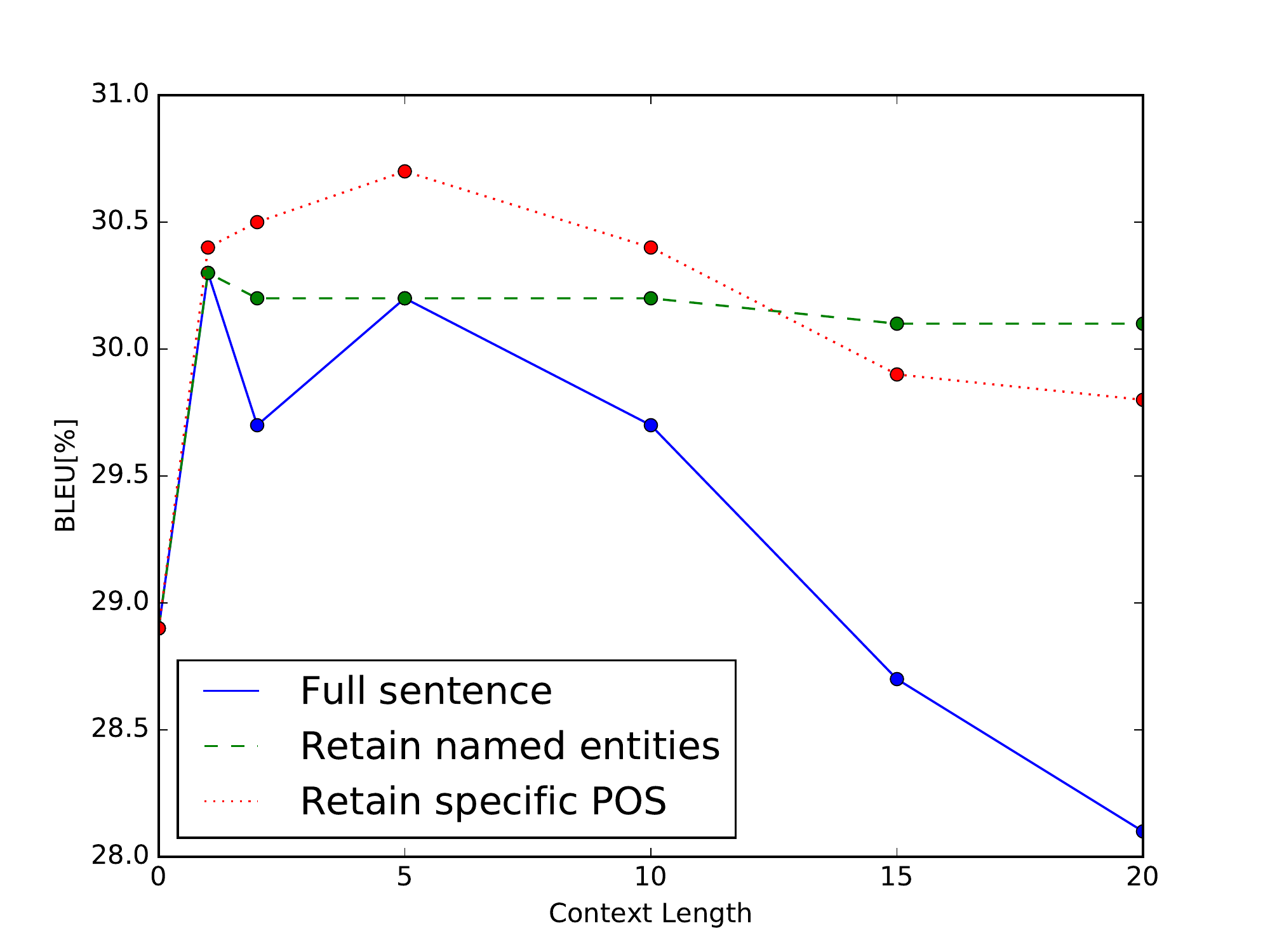}
\caption{WMT \textbf{en-de}}
\end{subfigure}
\caption{Translation performance as a function of document-level context length (in the number of sentences).}
\label{fig:length}
\end{figure*}

\subsection{Filtering Words in the Context}
\label{ssec:summary-exp}

To make the context modeling even lighter, we analyze the effectiveness of the filtered context (Section \ref{ssec:summary-theory}) in Table \ref{tab:summarization}.
All filtering methods shrink the context input drastically without a significant loss of performance.
Each method has its own motivation to retain only useful tokens in the context; the results show that they are all reasonable in practice.
In particular, using only named entities as context input, we achieve the same level of improvement with only 13\% of tokens in the full context sentences.
By filtering words in the context sentences, we can use more examples in each batch for a robust training.

\subsection{Context Length}
\label{ssec:length}

Filtered context inputs (Section \ref{ssec:summary-exp}) with a minimal encoding (Section \ref{ssec:architecture}) make it also feasible to include much longer context without much difficulty.
Most of previous works on document-level NMT have not examined context inputs longer than three sentences.

Figure \ref{fig:length} shows the translation performance with an increasing number of context sentences.
If we concatenate full context sentences (plain curves), the performance deteriorates severely.
We found that it is hard to fit such long sequences in memory as the training becomes very erratic.

The training is much more stable with filtered context; the dashed/dotted curves do not drop significantly even when using 20 context sentences.
In the English$\to$Italian task, the performance slightly improves up to 15 context sentences.
In the English$\to$German task, there is no improvement by extending the context length over 5 sentences.
This discrepancy can be explained with document lengths in each dataset (Table \ref{tab:corpus}).
The TED talk corpus for English$\to$Italian has much longer documents, thus it is probable to benefit from larger context windows.
However, in general we observe only marginal improvements by enlarging the context length to more than one sentence, as seen also in \newcite{bawden2018evaluating}, \newcite{miculicich2018document}, or \newcite{zhang2018improving}.

\section{Analysis}
\label{sec:analysis}

Simplifying the context encoder (Section \ref{ssec:architecture}) and filtering the context input (Section \ref{ssec:summary-exp}) are both inspired by the intuition that only a small part of the context is useful for NMT.
In order to verify this intuition rigorously, we conduct an extensive analysis on how document-level context helps the translation process, manually checking every output of sentence-level/document-level NMT models; automatic metrics are inherently not suitable for distinguishing document-level behavior.
Our analysis is not constrained to certain discourse phenomena which are favored in evaluating document-level models.
We quantify various causes of the improvements 1) regardless of its linguistic interpretability and 2) in a realistic scenario where not all the test examples require document-level context.
Here are the steps we take:
\begin{enumerate}\itemsep0em
    \item Translate a test set with a sentence-level baseline and a document-level model.
    \item Compute per-sentence {\ter} scores of outputs from both models.
    \item Select those cases where the document-level model improves the per-sentence {\ter} over the sentence-level baseline.
    \item Examine each case of 3 by looking at:
    \begin{itemize}
        \item Source, context, and translation outputs
        \item Attention distribution over the context tokens for each target token: averaged over all decoder layers/heads
        \item Gating activation (Equation \ref{eq:gating})
    \end{itemize}
    \item Classify each case into ``coreference'', ``topic-aware lexical choice'', or ``not interpretable''.
\end{enumerate}

Statistics of each category on the test sets are reported in Table \ref{tab:analysis}.
The manual inspection of translation outputs is done by a native-level speaker of Italian or German, respectively.

\begin{figure*}
    \centering
    \captionsetup[subfigure]{oneside,margin={1cm,0cm}}
    \begin{subfigure}[t]{0.32\linewidth}
    \includegraphics[width=\linewidth]{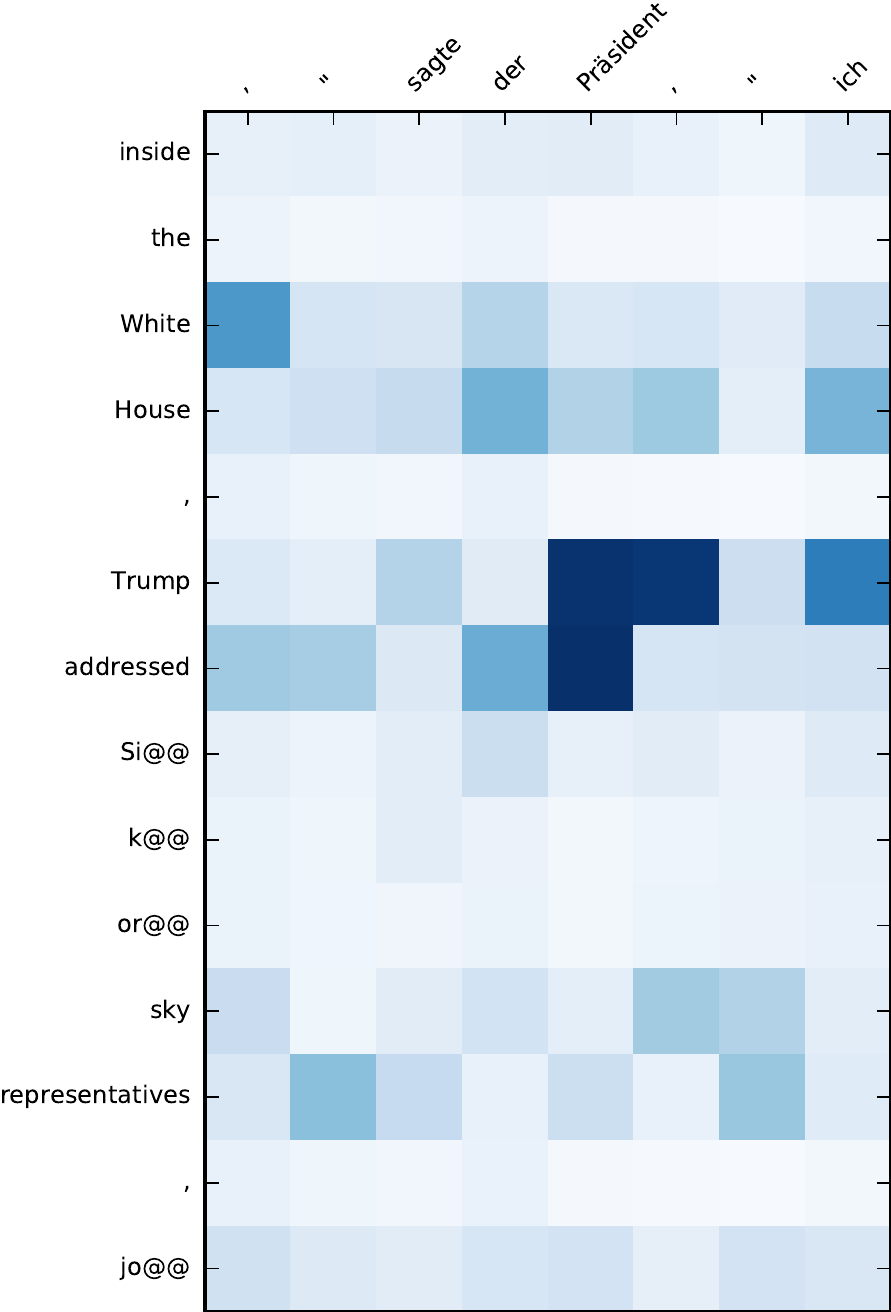}
    \caption{Coreference}
    \label{fig:analysis-co}
    \end{subfigure}
    \begin{subfigure}[t]{0.33\linewidth}
    \includegraphics[width=\linewidth]{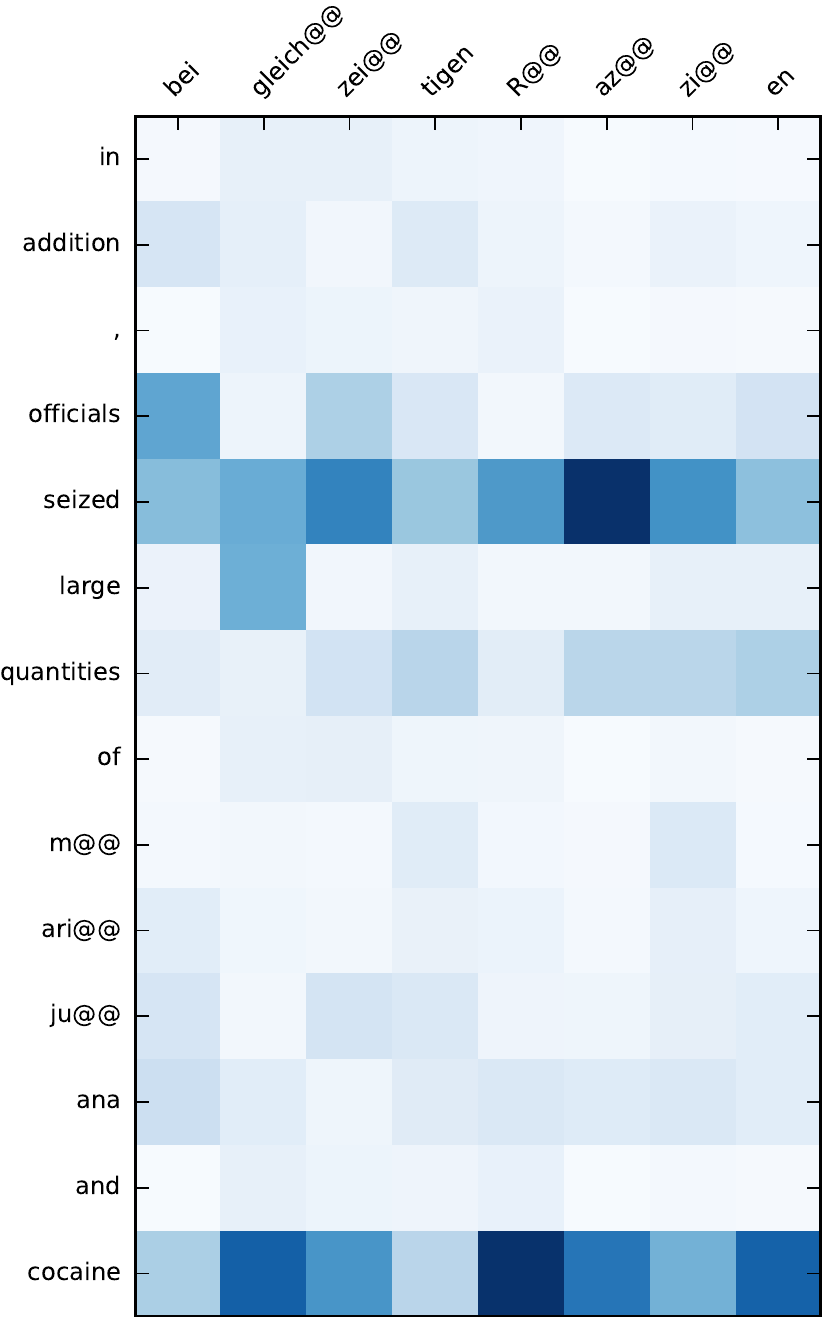}
    \caption{Topic-aware lexical choice}
    \label{fig:analysis-to}
    \end{subfigure}
    \begin{subfigure}[t]{0.33\linewidth}
    \includegraphics[width=\linewidth]{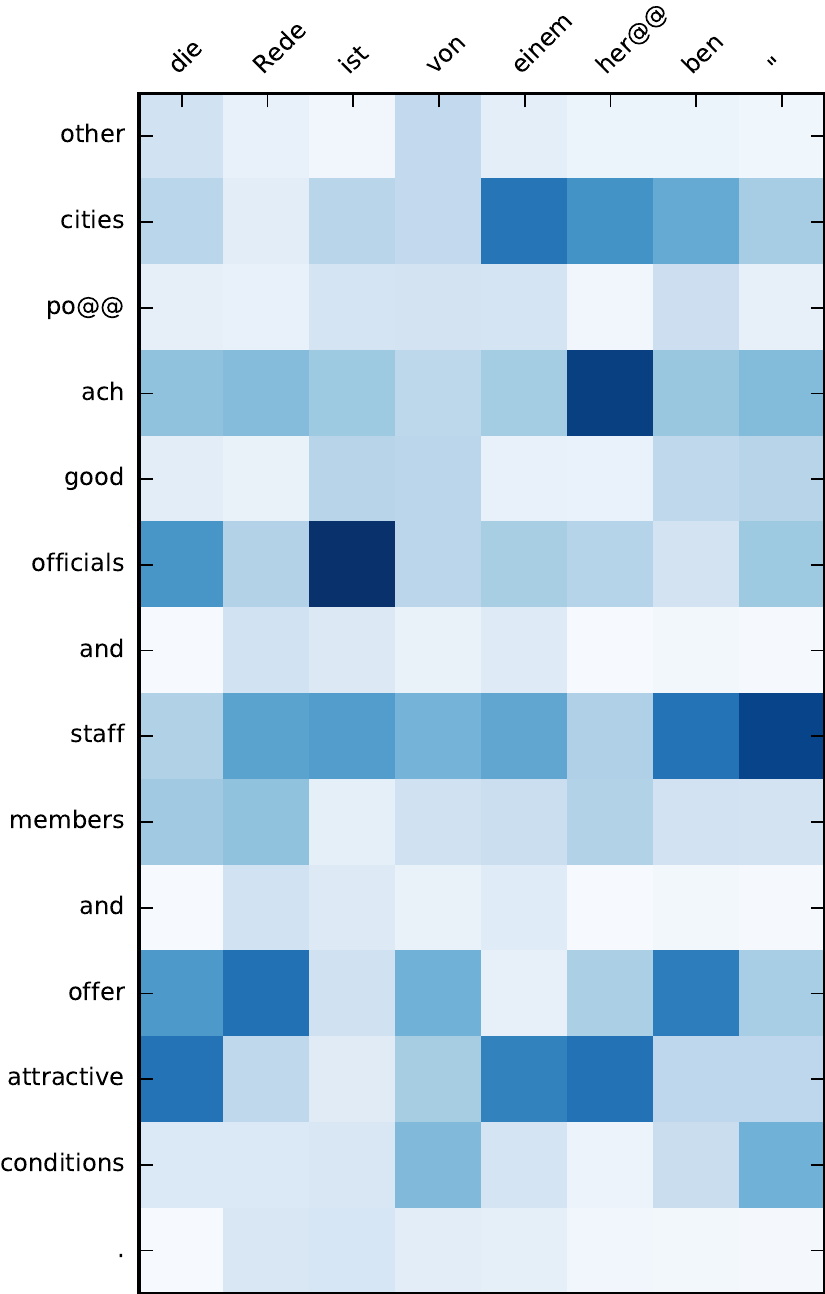}
    \caption{Not interpretable}
    \label{fig:analysis-un}
    \end{subfigure}
    \caption{Attention distribution over context words from target hypothesis.}
    \label{fig:analysis-ex}
\end{figure*}

Only a couple of cases belong to coreference, which is ironically the most advocated improvement in the literature on document-level NMT.
One of them is shown in Table \ref{tab:analysis-co}.
In the document-level NMT, the English word ``said'' is translated to a correct conjugation of ``sagen'' (= say) for the third person noun ``der Pr\"{a}sident'' (= the President).
This can be explained by the high attention energy on ``Trump'' (Figure \ref{fig:analysis-co}) in the context sentence.

Another interpretable cause is topic-aware lexical choice (Table \ref{tab:analysis-to}).
The document-level model actively attends to ``seized'' and ``cocaine'' in the context sentence (Figure \ref{fig:analysis-to}), and does not miss the source word ``raids'' in the translation (``Razzien'').
When it corrects the translation of polysemous words, it is related to word sense disambiguation \cite{gonzales2017improving,marvin2018exploring,pu2018integrating}.
This category includes also a coherence of text style in the translation outputs, depending on the context topic.

\begin{table}[!t]
    \centering
    \begin{tabular}{lcrr}
        \toprule
        & & \multicolumn{2}{c}{\#cases}\\
        \cmidrule{3-4}
        Category & & \multicolumn{1}{c}{\textbf{en-it}} & \multicolumn{1}{c}{\textbf{en-de}}\\
        \midrule
        Coreference & & 21 & 2 \\
        Topic-aware lexical choice & & 66 & 33 \\
        Not interpretable & & 292 & 1,211\\
        \midrule
        Total {\ter} improved & & 379 & 1,246\\
        \midrule
        Total & & 1,147 & 2,998\\
        \bottomrule
    \end{tabular}
    \caption{Causes of improvements by document-level context.}
    \label{tab:analysis}
\end{table}

We found that only 7.5\% of the {\ter}-improved cases can be interpreted as utilizing document-level context.
The other cases are mostly general improvements in adequacy or fluency which are not related to the given context.
Table \ref{tab:analysis-un} shows such an example.
It improves the translation by a long-range reordering and rephrasing some nouns, whose clues do not exist in the previous source sentence.
Its attention distribution over the context words is totally random and blurry (Figure \ref{fig:analysis-un}).

\begin{figure}[!hb]
    \centering
    \includegraphics[width=1.05\linewidth]{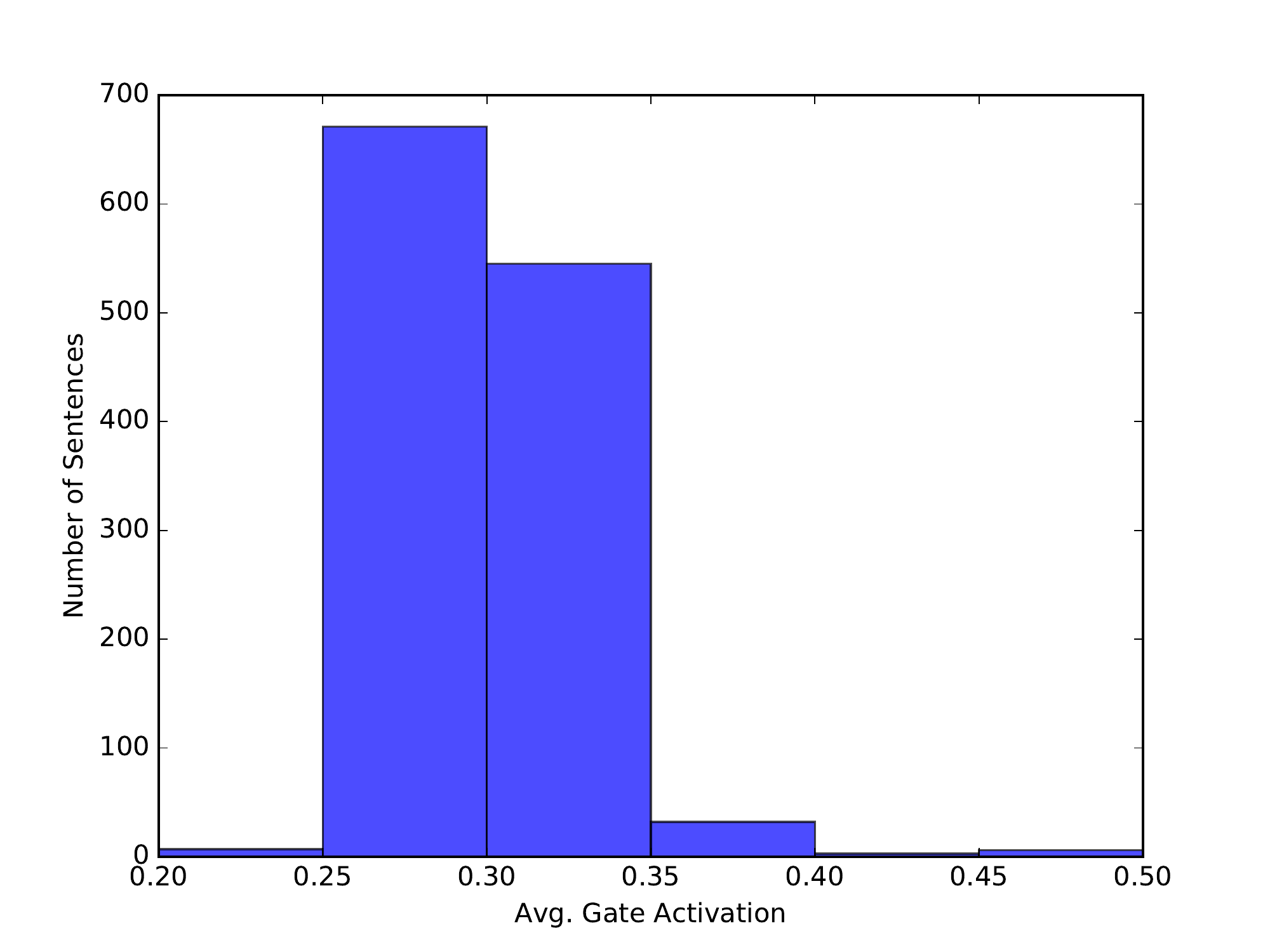}
    \caption{Gating activation for all {\ter}-improved cases of the English$\to$German task, averaged over all layers and target positions.}
    \label{fig:gating}\vspace{-1em}
\end{figure}

A possible reason for the non-interpretable improvements is regularization of the model, since the training data of our experiments are relatively small.
Figure \ref{fig:gating} shows that, for most of the improved cases, the model has non-negligible gating activation towards document-level context, even if the output seems not to benefit from the context.
It means that, when combining the encoded representations of context/current sentences, the model can reserve some of its capacity to the information from context inputs.
This might effectively mitigate overfitting to the given training data.

\FloatBarrier\clearpage
\begin{table*}[t!]
    \centering
    \begin{subfigure}[t]{\linewidth}
    \centering
	\begin{tabular}{lp{13cm}}
	\toprule
    \textbf{Previous src} & inside the White House, \underline{Trump} addressed Sikorsky representatives, joking with the media about his own fleet of company products.\\
    \textbf{Current src} & ``I know Sikorsky very well,'' the President said, ``I have three of them.''\\
    \midrule
    \textbf{Reference} & \glqq ich kenne Sikorsky sehr gut\grqq, sagte der Pr\"{a}sident, \glqq ich habe drei davon.\grqq \\
    \textbf{Sent-level hyp} & \glqq ich kenne Sikorsky sehr gut\grqq , \textbf{so} der Pr\"{a}sident, \glqq habe drei davon.\grqq\\
    \textbf{Doc-level hyp}& \glqq ich kenne Sikorsky sehr gut,\grqq \textbf{ sagte} der Pr\"{a}sident, \glqq ich habe drei davon\grqq .\\
    \bottomrule
	\end{tabular}
	\caption{Coreference}
	\label{tab:analysis-co}
    \end{subfigure}\vspace{0.6em}
    \begin{subfigure}[t]{\linewidth}
    \centering
	\begin{tabular}{lp{13cm}}
	\toprule
    \textbf{Previous src} & in addition, officials \underline{seized} large quantities of marijuana and \underline{cocaine}, firearms and several hundred thousand euros.\\
    \textbf{Current src} & at simultaneous raids in Italy, two people were detained.\\
    \midrule
    \textbf{Reference} & bei zeitgleichen Razzien in Italien wurden zwei Personen festgenommen.\\
    \textbf{Sent-level hyp} & gleichzeitig wurden in Italien zwei Personen verhaftet.\\
    \textbf{Doc-level hyp} & bei gleichzeitigen \textbf{Razzien} in Italien wurden zwei Menschen inhaftiert.\\ 
    \bottomrule
	\end{tabular}
	\caption{Topic-aware lexical choice}
	\label{tab:analysis-to}
    \end{subfigure}\vspace{0.6em}
    \begin{subfigure}[t]{\linewidth}
    \centering
	\begin{tabular}{lp{13cm}}
	\toprule
    \textbf{Previous src} & other cities poach good officials and staff members and offer attractive conditions.\\
    \textbf{Current src} & the talk is of a downright ``contest between public employers''.\\
    \midrule
    \textbf{Reference} & die Rede ist von einem regelrechten \glqq Wettbewerb der \"{o}ffentlichen Arbeitgeber\grqq.\\
    \textbf{Sent-level hyp} & das Gerede über einen \glqq Wettkampf zwischen \"{o}ffentlichen Arbeitgebern\grqq$\:$ ist von einem Gerechtigkeitstreit.\\
    \textbf{Doc-level hyp} & die Rede ist von einem herben \glqq Wettbewerb zwischen \"{o}ffentlichen Arbeitgebern\grqq.\\
    \bottomrule
	\end{tabular}
	\caption{Not interpretable}
	\label{tab:analysis-un}
    \end{subfigure}
    \caption{Example translation outputs for each analysis category (WMT English$\to$German newstest2018).}
    \label{tab:analysis-ex}
\end{table*}
\FloatBarrier

\begin{table}[t!]
    \centering
    \begin{tabular}{clccc}
        \toprule
        & & & \multicolumn{2}{c}{\textsc{Bleu} [\%]}\\
         \cmidrule{4-5}
        Condition & System & & \textbf{en-it} & \textbf{en-de}\\
        \midrule
        \multirow{2}{*}{\small{Dropout 0.1}} &
        Sentence-level & & 31.4 & 28.9\\
        & Document-level & & \textbf{32.5} & \textbf{30.3}\\
        \midrule
        \multirow{2}{*}{\small{Dropout 0.3}} &
        Sentence-level & & \textbf{33.7} & \textbf{32.3}\\
        & Document-level & & 33.5 & 32.0\\
        \midrule
        \multirow{2}{*}{\small{Large training data}} &
        Sentence-level & & - & \textbf{40.2}\\
        & Document-level & & - & 39.9\\
        \bottomrule
    \end{tabular}
    \caption{Sentence-level vs. document-level translation performance in different data/training conditions.}
    \label{tab:regularized}
\end{table}

We argue that the linguistic improvements with document-level NMT have been sometimes oversold, and the document-level components should be tested on top of a well-regularized NMT system.
In our experiments, we obtain a much stronger sentence-level baseline by applying a simple regularization (dropout), which the document-level model cannot outperform (Table \ref{tab:regularized}).

On a larger scale, we also built a sentence-level model with all parallel training data available for the WMT 2019 task and fine-tuned only with document-level data (Europarl, News Commentary, newstest2008-2014/2016).
The document-level training does not give any improvement in {\bleu} (last two rows of Table \ref{tab:regularized}).
There may exist document-level improvements which are not highlighted by the automatic metrics, but the amount of such improvements must be very small without a clear gain in {\bleu} or {\ter}.

\section{Conclusion}
\label{sec:conclusion}

In this work, we critically investigate the advantages of document-level NMT with a thorough qualitative analysis and expose the limit of its improvements in terms of context length and model complexity.
Regarding the questions asked in Section \ref{sec:introduction}, our answers are:
\begin{itemize}\itemsep0em
    \item In general, document-level context is utilized rarely in an interpretable way.
    \item We conjecture that a dominant cause of the improvements by document-level NMT is actually the regularization of the model.
    \item Not all of the words in the context are used in the model; we leave out redundant tokens without loss of performance.
    \item A long-range context gives only marginal additional improvements.
    \item Word embeddings are sufficient to model document-level context.
\end{itemize}

For a fair evaluation of document-level NMT methods, we argue that one should make a sentence-level NMT baseline as strong as possible first, i.e. by using more data or applying proper regularization.
This will get rid of by-product improvements from additional information flows and help to focus only on document-level errors in translation.
In this condition, we show that document-level NMT can barely improve translation metric scores against such strong baselines.
Targeted test sets \cite{bawden2018evaluating,voita2019when} might be helpful here to emphasize the document-level improvements.
However, one should bear in mind that a big improvement in such test sets may not carry over to practical scenarios with general test sets, where the number of document-level errors in translation is inherently small.

Given these conclusions, a future research direction would be building a lightweight post-editing model to correct only document-level errors, not complicating the sentence-level model too much for a very limited amount of document-level improvements.
To strengthen our arguments, we also plan to conduct the same qualitative analysis on other types of context inputs (e.g. translation history) and different domains.

Our implementation of document-level NMT methods is publicly available on the web.\footnote{\url{https://github.com/ducthanhtran/sockeye_document_context}}

\section*{Acknowledgments}

\begin{wrapfigure}{l}{0.2\textwidth}
\centering\vspace{-0.6em}
\includegraphics[width=0.22\textwidth]{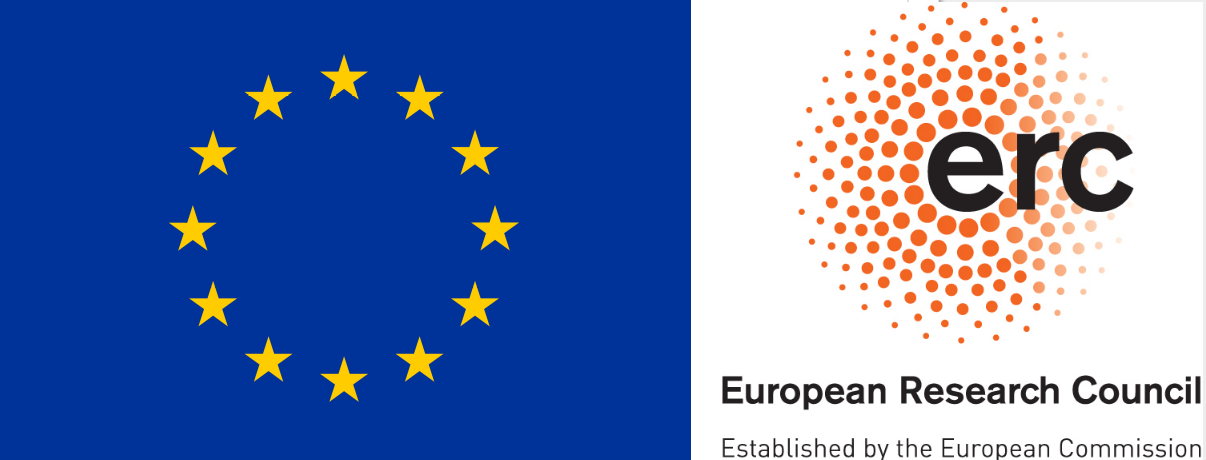}\vspace{-1em}
\end{wrapfigure}
This work has received funding from the European Research Council (ERC) (under the European Union's Horizon 2020 research and innovation programme, grant agreement No 694537, project ``SEQCLAS''). The work reflects only the authors' views and none of the funding agencies is responsible for any use that may be made of the information it contains.

The authors thank Tina Raissi for analyzing English$\rightarrow$Italian translations.

\bibliographystyle{acl_natbib}
\bibliography{references}

\appendix
\begin{table*}[!h]
	\begin{tabular}{l}
	\hspace{-0.2cm}\large\textbf{A\hspace{0.4cm}Context Word Filtering Details}\\ \\
	\hspace{-0.2cm}The tables below follow the tagging conventions of \textsc{Flair} (\small{https://github.com/zalandoresearch/flair}).
    \end{tabular}
    \vspace{-2cm}
\end{table*}
\begin{table*}[!h]
	\begin{center}
		\begin{tabular}{l|p{5.5cm}||l|p{4cm}}
			Tag & Examples & Tag & Examples \\
			\toprule
			Location & mountain ranges, bodies of water & Law & named laws \\
			Organization & companies, agencies & Language & any named language \\
			Person & People, fictional characters & Dates & important holidays \\
			NORP & nationalities, political groups & Time & times smaller than a day \\
			Facility & buildings, airports & Percent & percentage values \\
			GPE & countries, cities & Money & monetary values \\
			Product & vehicles, food & Ordinal & first, second, $\ldots$ \\
			Event & named hurricanes, battles & Cardinal & other numeric values
		\end{tabular}
	\end{center}
	\caption{Retained named entities.}\label{tab:retainedTags:ner}
\end{table*}

\begin{table*}[!h]
	\begin{center}
		\begin{tabular}{l|p{4cm}||l|p{6cm}}
			Tag & Description & Tag & Description\\
			\toprule
			CC & Coordinating conjunction & PDT & Predeterminer\\
			CD & Cardinal number & PRP & Personal pronoun\\
			CODE & Code IDs & PRP\$ & Possessive pronoun\\
			DT & Determiner & UH & Interjection \\
			FW & Foreign word & VBG & Verb, gerund/present particle\\
			MD & Modal & VBN & Verb, past participle\\
			NN & Noun, singular & VBP & Verb, non-3rd person singular present\\
			NNP & Proper noun, singular & VBZ & Verb, 3rd person singular present\\
			NNS & Noun, plural & VB & Verb, base form \\
			NNPS & Proper noun, plural & VBD & Verb, past tense \\
		\end{tabular}
	\end{center}
	\caption{Retained parts-of-speech.}\label{tab:retainedTags:pos}
\end{table*}

\end{document}